\documentclass[11pt,a4paper]{article}
\usepackage[hyperref]{emnlp2018}
\usepackage{times}
\usepackage{latexsym}
\usepackage{multirow}
\usepackage{mathtools}
\usepackage{floatrow}
\usepackage{subcaption}
\usepackage{url}

\aclfinalcopy 

\title{Linguistically-Informed Self-Attention for Semantic Role Labeling}

\author{Emma Strubell$^1$, Patrick Verga$^1$, Daniel Andor$^2$, David Weiss$^2$ and Andrew McCallum$^1$ \\
  $^1$College of Information and Computer Sciences \\
    University of Massachusetts Amherst \\
    {\tt \{strubell, pat, mccallum\}@cs.umass.edu} \\
  $^2$Google AI Language \\
  New York, NY \\
  {\tt\{andor, djweiss\}@google.com}}

\date{}

\begin{document}
\maketitle
\begin{abstract}
Current state-of-the-art semantic role labeling (SRL) uses a deep neural network with no explicit linguistic features. 
However, prior work has shown that gold syntax trees can dramatically improve SRL decoding, suggesting the possibility of increased accuracy from explicit modeling of syntax.
In this work, we present linguistically-informed self-attention (LISA): a neural network model that combines multi-head self-attention with multi-task learning across dependency parsing, part-of-speech tagging, predicate detection and SRL.  Unlike previous models which require significant pre-processing to prepare linguistic features, LISA can incorporate syntax using merely raw tokens as input, encoding the sequence only once to simultaneously perform parsing, predicate detection and role labeling for all predicates. Syntax is incorporated by training one attention head to attend to syntactic parents for each token. Moreover, if a high-quality syntactic parse is already available, it can be beneficially injected at test time without re-training our SRL model.
In experiments on CoNLL-2005 SRL, LISA achieves new state-of-the-art performance for a model using predicted predicates and standard word embeddings, attaining 2.5 F1 absolute higher than the previous state-of-the-art on newswire and more than 3.5 F1 on out-of-domain data, nearly 10\% reduction in error. On ConLL-2012 English SRL we also show an improvement of more than 2.5 F1. LISA also out-performs the state-of-the-art with contextually-encoded (ELMo) word representations, by nearly 1.0 F1 on news and more than 2.0 F1 on out-of-domain text.
\end{abstract}

\section{Introduction}
Semantic role labeling (SRL) extracts a high-level representation of meaning from a sentence, labeling e.g.\ \emph{who} did \emph{what} to \emph{whom}. Explicit representations of such semantic information have been shown to improve results in challenging downstream tasks such as dialog systems \citep{tur2005semi,chen2013unsupervised}, machine reading \citep{berant2014modeling, wang2015machine} and translation \citep{liu2010semantic,bazrafshan2013semantic}. 

Though syntax was long considered an obvious prerequisite for SRL systems \citep{levin1993english,punyakanok2008importance}, recently deep neural network architectures have surpassed syntactically-informed models \citep{zhou2015end, marcheggiani2017simple, he2017deep, tan2018deep, he2018jointly}, achieving state-of-the art SRL performance with no explicit modeling of syntax. 
An additional benefit of these end-to-end models is that they require just raw tokens and (usually) detected predicates as input, whereas richer linguistic features typically require extraction by an auxiliary pipeline of models.

Still, recent work \citep{roth2016neural,he2017deep,marcheggiani2017encoding} indicates that neural network models could see even higher accuracy gains by leveraging syntactic information rather than ignoring it. \citet{he2017deep} indicate that many of the errors made by a syntax-free neural network on SRL are tied to certain syntactic confusions such as prepositional phrase attachment, and show that while constrained inference using a relatively low-accuracy predicted parse can provide small improvements in SRL accuracy, providing a gold-quality parse leads to substantial gains. \citet{marcheggiani2017encoding} incorporate syntax from a high-quality parser \citep{kiperwasser2016simple} using graph convolutional neural networks \citep{kipf2017semi}, but like \citet{he2017deep} they attain only small increases over a model with no syntactic parse, and even perform worse than a syntax-free model on out-of-domain data. These works suggest that though syntax has the potential to improve neural network SRL models, we have not yet designed an architecture which maximizes the benefits of auxiliary syntactic information. 

In response, we propose \emph{linguistically-informed self-attention} (LISA): a model that combines multi-task learning \citep{caruana1993multitask} with stacked layers of multi-head self-attention \citep{vaswani2017attention}; the model is trained to: (1) jointly predict parts of speech and predicates; (2) perform parsing; and (3) attend to syntactic parse parents, while (4) assigning semantic role labels. Whereas prior work typically requires separate models to provide linguistic analysis, including most syntax-free neural models which still rely on external predicate detection, our model is truly end-to-end: earlier layers are trained to predict prerequisite parts-of-speech and predicates, the latter of which are supplied to later layers for scoring. Though prior work re-encodes each sentence to predict each desired task and again with respect to each predicate to perform SRL,
we more efficiently encode each sentence only once, predict its predicates, part-of-speech tags and labeled syntactic parse, then predict the semantic roles for all predicates in the sentence in parallel. The model is trained such that, as syntactic parsing models improve, providing high-quality parses at test time will improve its performance, allowing the model to leverage updated parsing models without requiring re-training. 

In experiments on the CoNLL-2005 and CoNLL-2012 datasets we show that our linguistically-informed models out-perform the syntax-free state-of-the-art. On CoNLL-2005 with predicted predicates and standard word embeddings, our single model out-performs the previous state-of-the-art model on the WSJ test set by 2.5 F1 points absolute. On the challenging out-of-domain Brown test set, our model improves substantially over the previous state-of-the-art by more than 3.5 F1, a nearly 10\% reduction in error. On CoNLL-2012, our model gains more than 2.5 F1 absolute over the previous state-of-the-art. Our models also show improvements when using contextually-encoded word representations \citep{peters2018deep}, obtaining nearly 1.0 F1 higher than the state-of-the-art on CoNLL-2005 news and more than 2.0 F1 improvement on out-of-domain text.\footnote{Our implementation in TensorFlow \citep{abadi2015tensorflow} is available at : \protect\url{http://github.com/strubell/LISA}}

\section{Model}

\begin{figure}[t]
\begin{center}
\includegraphics[scale=.8]{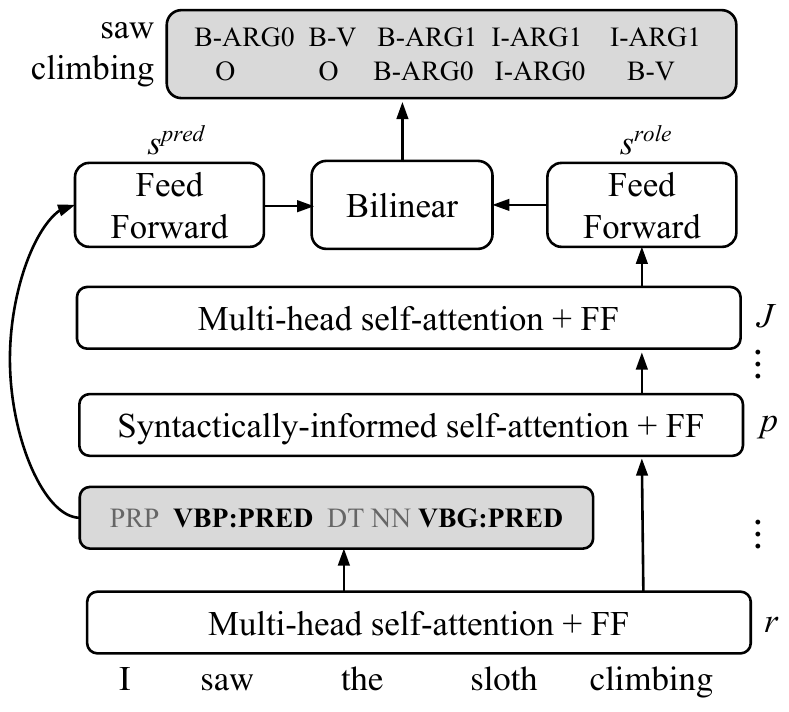}
\caption{Word embeddings are input to $J$ layers of multi-head self-attention. In layer $p$ one attention head is trained to attend to parse parents (Figure \ref{attention-fig}). Layer $r$ is input for a joint predicate/POS classifier. Representations from layer $r$ corresponding to predicted predicates are passed to  a bilinear operation scoring distinct predicate and role representations to produce per-token SRL predictions with respect to each predicted predicate.\label{architecture-fig}}
\end{center}
\end{figure}

\begin{figure}[t]
\begin{center}
\includegraphics[scale=.24]{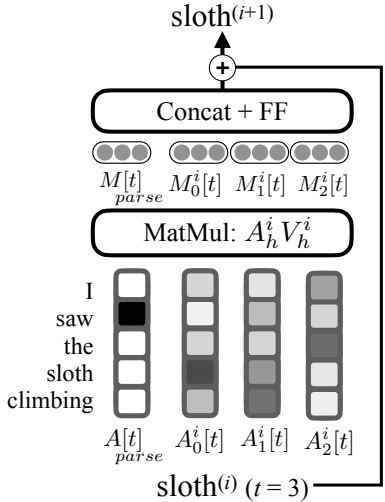}
\caption{Syntactically-informed self-attention for the query word \emph{sloth}. Attention weights $A_{parse}$ heavily weight the token's syntactic governor, \emph{saw}, in a weighted average over the token values $V_{parse}$. The other attention heads act as usual, and the attended representations from all heads are concatenated and projected through a feed-forward layer to produce the syntactically-informed representation for \emph{sloth}. \label{attention-fig}}
\end{center}
\end{figure}

Our goal is to design an efficient neural network model which makes use of linguistic information as effectively as possible in order to perform end-to-end SRL. LISA achieves this by combining: (1) A new technique of supervising neural attention to predict syntactic dependencies with (2) multi-task learning across four related tasks. 

Figure~\ref{architecture-fig} depicts the overall architecture of our model. The basis for our model is the Transformer encoder introduced by \citet{vaswani2017attention}: we transform word embeddings into contextually-encoded token representations using stacked multi-head self-attention and feed-forward layers (\S\ref{sec:self-attn}). 

To incorporate syntax, one self-attention head is trained to attend to each token's syntactic parent, allowing the model to use this attention head as an oracle for syntactic dependencies. We introduce this \emph{syntactically-informed self-attention} (Figure \ref{attention-fig}) in more detail in \S\ref{sec:syntax-attn}. 

Our model is designed for the more realistic setting in which gold predicates are not provided at test-time. Our model predicts predicates and integrates part-of-speech (POS) information into earlier layers by re-purposing representations closer to the input to predict predicate and POS tags using hard parameter sharing (\S\ref{sec:MTL}). We simplify optimization and benefit from shared statistical strength derived from highly correlated POS and predicates by treating tagging and predicate detection as a single task, performing multi-class classification into the joint Cartesian product space of POS and predicate labels. 

Though typical models, which re-encode the sentence for each predicate, can simplify SRL to token-wise tagging, our joint model requires a different approach to classify roles with respect to each predicate. Contextually encoded tokens are projected to distinct \emph{predicate} and \emph{role} embeddings (\S\ref{sec:srl}), and each predicted predicate is scored with the sequence's role representations using a bilinear model (Eqn.~\ref{eqn:bilinear}), producing per-label scores for BIO-encoded semantic role labels for each token and each semantic frame. 

The model is trained end-to-end by maximum likelihood using stochastic gradient descent (\S\ref{sec:train-opt}). 

\subsection{Self-attention token encoder \label{sec:self-attn}}
The basis for our model is a multi-head self-attention token encoder, recently shown to achieve state-of-the-art performance on SRL \citep{tan2018deep}, and which provides a natural mechanism for incorporating syntax, as described in \S\ref{sec:syntax-attn}. Our implementation replicates \citet{vaswani2017attention}. 

The input to the network is a sequence $\mathcal{X}$ of $T$ token representations $x_t$. In the standard setting these token representations are initialized to pre-trained word embeddings, but we also experiment with supplying pre-trained ELMo representations combined with task-specific learned parameters, which have been shown to substantially improve performance of other SRL models \citep{peters2018deep}.
For experiments with gold predicates, we concatenate a predicate indicator embedding $p_t$ following previous work \citep{he2017deep}.

We project\footnote{All linear projections include bias terms, which we omit in this exposition for the sake of clarity.} these input embeddings to a representation that is the same size as the output of the self-attention layers. We then add a positional encoding vector computed as a deterministic sinusoidal function of $t$, since the self-attention has no innate notion of token position. 

We feed this token representation as input to a series of $J$ residual multi-head self-attention layers with feed-forward connections. Denoting the $j$th self-attention layer as $T^{(j)}(\cdot)$, the output of that layer $s_t^{(j)}$, and $LN(\cdot)$ layer normalization, the following recurrence applied to initial input $c_t^{(p)}$:
\begin{align}
\label{eqn:overall}
s_t^{(j)} = LN(s_t^{(j-1)} + T^{(j)}(s_t^{(j-1)}))
\end{align}
gives our final token representations $s_t^{(j)}$. Each $T^{(j)}(\cdot)$ consists of: (a) multi-head self-attention and (b) a feed-forward projection.

The multi-head self attention consists of $H$ attention heads, each of which learns a distinct attention function to attend to all of the tokens in the sequence. This self-attention is performed for each token for each head, and the results of the $H$ self-attentions are concatenated to form the final self-attended representation for each token. 

Specifically, consider the matrix $S^{(j-1)}$ of $T$ token representations at layer $j-1$. For each attention head $h$, we project this matrix into distinct key, value and query representations $K_h^{(j)}$, $V_h^{(j)}$ and $Q_h^{(j)}$ of dimensions $T\times d_k$, $T\times d_q$, and $T\times d_v$, respectively. We can then multiply $Q_h^{(j)}$ by $K_h^{(j)}$ to obtain a $T\times T$ matrix of attention weights $A_h^{(j)}$ between each pair of tokens in the sentence. Following \citet{vaswani2017attention} we perform scaled dot-product attention: We scale the weights by the inverse square root of their embedding dimension and normalize with the softmax function to produce a distinct distribution for each token over all the tokens in the sentence:
\begin{align}
A_h^{(j)} = \mathrm{softmax}(d_{k}^{-0.5}Q_h^{(j)}{K_h^{(j)}}^T)
\end{align}
These attention weights are then multiplied by $V_h^{(j)}$ for each token to obtain the self-attended token representations $M_h^{(j)}$:
\begin{align}
M_h^{(j)} = A_h^{(j)}V_h^{(j)}
\end{align}
Row $t$ of $M_h^{(j)}$, the self-attended representation for token $t$ at layer $j$, is thus the weighted sum with respect to $t$ (with weights given by $A_h^{(j)}$) over the token representations in $V_h^{(j)}$. 

The outputs of all attention heads for each token are concatenated, and this representation is passed to the feed-forward layer, which consists of two linear projections each followed by leaky ReLU activations \citep{maas2012rectifier}. We add the output of the feed-forward to the initial representation and apply layer normalization to give the final output of self-attention layer $j$, as in Eqn. \ref{eqn:overall}.

\subsection{Syntactically-informed self-attention \label{sec:syntax-attn}}
Typically, neural attention mechanisms are left on their own to learn to attend to relevant inputs. Instead, we propose training the self-attention to attend to specific tokens corresponding to the syntactic structure of the sentence as a mechanism for passing linguistic knowledge to later layers. 

Specifically, we replace one attention head with the deep bi-affine model of \citet{dozat2017deep}, trained to predict syntactic dependencies. Let $A_{parse}$ be the parse attention weights, at layer $i$. Its input is the matrix of token representations $S^{(i-1)}$. As with the other attention heads, we project $S^{(i-1)}$ into key, value and query representations, denoted $K_{parse}$, $Q_{parse}$, $V_{parse}$. Here the key and query projections correspond to $parent$ and $dependent$ representations of the tokens, and we allow their dimensions to differ from the rest of the attention heads to more closely follow the implementation of \citet{dozat2017deep}. Unlike the other attention heads which use a dot product to score key-query pairs, we score the compatibility between $K_{parse}$ and $Q_{parse}$ using a bi-affine operator $U_{heads}$ to obtain attention weights:
\begin{align}
A_{parse} = \mathrm{softmax}(Q_{parse} U_{heads} K_{parse}^T)
\end{align}
These attention weights are used to compose a weighted average of the value representations $V_{parse}$ as in the other attention heads.

We apply auxiliary supervision at this attention head to encourage it to attend to each token's parent in a syntactic dependency tree, and to encode information about the token's dependency label. Denoting the attention weight from token $t$ to a candidate head $q$ as $A_{parse}[t,q]$, we model the probability of token $t$ having parent $q$ as:
\begin{align}
P(q=\mathrm{head}(t) \mid \mathcal{X}) = A_{parse}[t, q]
\end{align}
using the attention weights $A_{parse}[t]$ as the distribution over possible heads for token $t$. We define the root token as having a self-loop. This attention head thus emits a directed graph\footnote{Usually the head emits a tree, but we do not enforce it here.} where each token's parent is the token to which the attention $A_{parse}$ assigns the highest weight. 

We also predict dependency labels using per-class bi-affine operations between parent and dependent representations $Q_{parse}$ and $K_{parse}$ to produce per-label scores, with locally normalized probabilities over dependency labels $y_t^{dep}$ given by the softmax function. We refer the reader to \citet{dozat2017deep} for more details.

This attention head now becomes an oracle for syntax, denoted $\mathcal{P}$, providing a dependency parse to downstream layers. This model not only predicts its own dependency arcs, but allows for the injection of auxiliary parse information at test time by simply setting $A_{parse}$ to the parse parents produced by e.g.\ a state-of-the-art parser. In this way, our model can benefit from improved, external parsing models without re-training.
Unlike typical multi-task models, ours maintains the ability to leverage external syntactic information.

\subsection{Multi-task learning \label{sec:MTL}}
We also share the parameters of lower layers in our model to predict POS tags and predicates. Following \citet{he2017deep}, we focus on the end-to-end setting, 
where predicates must be predicted on-the-fly. Since we also train our model to predict syntactic dependencies, it is beneficial to give the model knowledge of POS information. While much previous work employs a pipelined approach to both POS tagging for dependency parsing and predicate detection for SRL, we take a multi-task learning (MTL) approach \citep{caruana1993multitask}, sharing the parameters of earlier layers in our SRL model with a joint POS and predicate detection objective. Since POS is a strong predictor of predicates\footnote{All predicates in CoNLL-2005 are verbs; CoNLL-2012 includes some nominal predicates.} 
and the complexity of training a multi-task model increases with the number of tasks, we combine POS tagging and predicate detection into a joint label space: For each POS tag {\sc tag} which is observed co-occurring with a predicate, we add a label of the form {\sc tag:predicate}.

Specifically, we feed the representation $s_t^{(r)}$ from a layer $r$ preceding the syntactically-informed layer $p$ to a linear classifier to produce per-class scores $r_t$ for token $t$. We compute locally-normalized probabilities using the softmax function: $P(y_t^{prp} \mid \mathcal{X}) \propto \exp(r_t)$, where $y_t^{prp}$ is a label in the joint space. 

\subsection{Predicting semantic roles \label{sec:srl}}
Our final goal is to predict semantic roles for each predicate in the sequence. We score each predicate
against each token in the sequence using a bilinear operation, producing per-label scores for each token for each predicate, with predicates and syntax determined by oracles $\mathcal{V}$ and $\mathcal{P}$. 

First, we project each token representation $s_t^{(J)}$ to a predicate-specific representation $s_t^{pred}$ and a role-specific representation $s_t^{role}$.
We then provide these representations to a bilinear transformation $U$ for scoring. So, the role label scores $s_{ft}$ for the token at index $t$ with respect to the predicate at index $f$ (i.e. token $t$ and frame $f$) are given by:
\begin{align}
\label{eqn:bilinear}
s_{ft} = (s_f^{pred})^T U s_t^{role}
\end{align}
which can be computed in parallel across all semantic frames in an entire minibatch. We calculate a locally normalized distribution over role labels for token $t$ in frame $f$ using the softmax function: $P(y_{ft}^{role}\mid \mathcal{P},\mathcal{V}, \mathcal{X}) \propto \exp(s_{ft})$.

At test time, we perform constrained decoding using the Viterbi algorithm to emit valid sequences of BIO tags, using unary scores $s_{ft}$ and the transition probabilities given by the training data.

\subsection{Training \label{sec:train-opt}}
We maximize the sum of the likelihoods of the individual tasks.
In order to maximize our model's ability to leverage syntax, during training we clamp $\mathcal{P}$ to the gold parse ($\mathcal{P}_G$) and $\mathcal{V}$ to gold predicates $\mathcal{V}_G$ when passing parse and predicate representations to later layers, whereas syntactic head prediction and joint predicate/POS prediction are conditioned only on the input sequence $\mathcal{X}$. The overall objective is thus:
\begin{align}
\frac{1}{T}\sum_{t=1}^T\Big[&\sum_{f=1}^F \log P(y_{ft}^{role}\mid \mathcal{P}_G, \mathcal{V}_G, \mathcal{X}) \nonumber \\ 
&+ \log P(y_t^{prp}\mid \mathcal{X}) \nonumber \\ 
&+ \lambda_1 \log P(\mathrm{head}(t)\mid \mathcal{X}) \nonumber \\ 
&+ \lambda_2 \log P(y_t^{dep} \mid \mathcal{P}_G, \mathcal{X}) \label{eqn:rel-term} \Big]
\end{align}
where $\lambda_1$ and $\lambda_2$ are penalties on the syntactic attention loss. 

We train the model using Nadam \citep{dozat2016incorporating} SGD combined with the learning rate schedule in \citet{vaswani2017attention}. In addition to MTL, we regularize our model using dropout \citep{srivastava2014dropout}. We use gradient clipping to avoid exploding gradients \citep{bengio1994learning, pascanu2013on}. Additional details on optimization and hyperparameters are included in Appendix \ref{sec:supplemental}.

\section{Related work}

Early approaches to SRL \citep{pradhan2005semantic,surdeanu2007combination,johansson2008dependency,toutanova2008global} focused on developing rich sets of linguistic features as input to a linear model, often combined with complex constrained inference e.g. with an ILP \citep{punyakanok2008importance}. \citet{tackstrom2015efficient} showed that constraints could be enforced more efficiently using a clever dynamic program for exact inference. 
\citet{sutton2005joint} modeled syntactic parsing and SRL jointly, and \citet{lewis2015joint} jointly modeled SRL and CCG parsing. 

\citet{collobert2011natural} were among the first to use a neural network model for SRL, a CNN over word embeddings which failed to out-perform non-neural models.
\citet{fitzgerald2015semantic} successfully employed neural networks by embedding lexicalized features and providing them as factors in the model of \citet{tackstrom2015efficient}.

More recent neural models are syntax-free. \citet{zhou2015end}, \citet{marcheggiani2017simple} and \citet{he2017deep} all use variants of deep LSTMs with constrained decoding, while \citet{tan2018deep} apply self-attention to obtain state-of-the-art SRL with gold predicates. Like this work, \citet{he2017deep} present end-to-end experiments, predicting predicates using an LSTM, and \citet{he2018jointly} jointly predict SRL spans and predicates in a model based on that of \citet{lee2017end}, obtaining state-of-the-art predicted predicate SRL. Concurrent to this work, \citet{peters2018deep} and \citet{he2018jointly} report significant gains on PropBank SRL by training a wide LSTM language model and using a task-specific transformation of its hidden representations (ELMo) as a deep, and computationally expensive, alternative to typical word embeddings. We find that LISA obtains further accuracy increases when provided with ELMo word representations, especially on out-of-domain data.

Some work has incorporated syntax into neural models for SRL. \citet{roth2016neural} incorporate syntax by embedding dependency paths, and similarly \citet{marcheggiani2017encoding} encode syntax using a graph CNN over a predicted syntax tree, out-performing models without syntax on CoNLL-2009. These works are limited to incorporating partial dependency paths between tokens whereas our technique incorporates the entire parse. Additionally, \citet{marcheggiani2017encoding} report that their model does not out-perform syntax-free models on out-of-domain data, a setting in which our technique excels.


MTL \citep{caruana1993multitask} is popular in NLP, and others have proposed MTL models which incorporate subsets of the tasks we do \citep{collobert2011natural, zhang2016stack, hashimoto2017joint, peng2017deep, swayamdipta2017}, and we build off work that investigates where and when to combine different tasks to achieve the best results \citep{sogaard2016deep, bingel2017identifying, alonso2017when}. Our specific method of incorporating supervision into self-attention is most similar to the concurrent work of \citet{liu2018learning}, who use edge marginals produced by the matrix-tree algorithm as attention weights for document classification and natural language inference.


The question of training on gold versus predicted labels is closely related to learning to search \citep{daume2009search,ross2011reduction,chang2015learning} and scheduled sampling \citep{bengio2015scheduled}, with applications in NLP to sequence labeling and transition-based parsing \citep{choi2011getting, goldberg2012dynamic,ballesteros2016training}. Our approach may be interpreted as an extension of teacher forcing \citep{williams1989learning} to MTL. We leave exploration of more advanced scheduled sampling techniques to future work. 

\section{Experimental results}

\begin{table*}[t!]
\begin{tabular}{llllllllllll}
& \multicolumn{3}{c}{Dev} && \multicolumn{3}{c}{WSJ Test} && \multicolumn{3}{c}{Brown Test} \\ \cline{2-4} \cline{6-8} \cline{10-12}
GloVe & P & R & F1 && P & R & F1 && P & R & F1\\ \hline \hline
\citet{he2017deep} PoE & 81.8 &  81.2 & 81.5 & & 82.0 & 83.4 & 82.7 && 69.7 &  70.5 & 70.1 \\ 
\citet{he2018jointly} & 81.3 & 81.9 & 81.6 & & 81.2 & 83.9 & 82.5 && 69.7 & 71.9 & 70.8\\ \hline
SA &  83.52 & 81.28 & 82.39 &&  84.17 &	83.28 &	83.72 && 72.98 & 70.1 & 71.51 \\ 
LISA &  83.1 & 81.39 &  82.24 && 84.07 & 83.16 & 83.61 && 73.32 & 70.56 & 71.91\\ 			
\ \ \ \ +D\&M & {\bf 84.59} & {\bf 82.59} &	{\bf 83.58} && {\bf 85.53} & {\bf 84.45} & {\bf 84.99} && {\bf 75.8} & {\bf 73.54} & {\bf 74.66}\\ 	
\ \ \ \ \emph{+Gold} & \emph{87.91} & \emph{85.73} & \emph{86.81} && --- & --- & --- && --- & --- & --- \\
& & & && & & && & & \\
ELMo & & & && & & && & & \\ \hline \hline
\citet{he2018jointly} & 84.9 & {\bf 85.7} & 85.3 & & 84.8 & {\bf 87.2} & 86.0 && 73.9 & {\bf 78.4} & 76.1\\ \hline
SA &  85.78	& 84.74	& 85.26 &&  86.21 &	85.98 &	86.09 && 77.1 &	75.61 &	76.35 \\ 
LISA &  {\bf 86.07} & 84.64 & {\bf 85.35} && 86.69 & 86.42 & 86.55 && 78.95 & 77.17 &	78.05\\ 			
\ \ \ \ +D\&M &85.83 &	84.51 &	85.17 && {\bf 87.13} & 86.67 & {\bf 86.90} && {\bf 79.02} & 77.49 & {\bf 78.25}\\ 	
\ \ \ \ \emph{+Gold} & \emph{88.51} & \emph{86.77} & \emph{87.63} && --- & --- & --- && --- & --- & ---
\end{tabular}
\caption{Precision, recall and F1 on the CoNLL-2005 development and test sets. \label{tab:conll05-results}}
\end{table*}

We present results on the CoNLL-2005 shared task \citep{carreras2005introduction} and the CoNLL-2012 English subset of OntoNotes 5.0 \citep{pradhan2013towards}, achieving state-of-the-art results for a single model with predicted predicates on both corpora. We experiment with both standard pre-trained GloVe word embeddings \citep{pennington2014glove} and pre-trained ELMo representations with fine-tuned task-specific parameters \citep{peters2018deep} in order to best compare to prior work. Hyperparameters that resulted in the best performance on the validation set were selected via a small grid search, and models were trained for a maximum of 4 days on one TitanX GPU using early stopping on the validation set.
We convert constituencies to dependencies using the Stanford head rules v3.5 \citep{deMarneffe2008}. A detailed description of hyperparameter settings and data pre-processing can be found in Appendix~\ref{sec:supplemental}.

We compare our {\bf LISA} models to four strong baselines: For experiments using predicted predicates, we compare to \citet{he2018jointly} and the ensemble model ({\bf PoE}) from \citet{he2017deep}, as well as a version of our own self-attention model which does not incorporate syntactic information ({\bf SA}). To compare to more prior work, we present additional results on CoNLL-2005 with models given gold predicates at test time. In these experiments we also compare to \citet{tan2018deep}, the previous state-of-the art SRL model using gold predicates and standard embeddings.

We demonstrate that our models benefit from injecting state-of-the-art predicted parses at test time ({\bf +D\&M}) by fixing the attention to parses predicted by \citet{dozat2017deep}, the winner of the 2017 CoNLL shared task \citep{zeman2017conll} which we re-train using ELMo embeddings. In all cases, using these parses at test time improves performance. 

We also evaluate our model using the gold syntactic parse at test time ({\bf +Gold}), to provide an upper bound for the benefit that syntax could have for SRL using LISA. These experiments show that despite LISA's strong performance, there remains substantial room for improvement. In \S\ref{sec:analysis} we perform further analysis comparing SRL models using gold and predicted parses.

\begin{table}
\begin{tabular}{llll}
WSJ Test & P & R & F1 \\ \hline \hline
\citet{he2018jointly} & 84.2 & 83.7 & 83.9 \\
\citet{tan2018deep} & 84.5 & 85.2 & 84.8 \\ \hline
SA & 84.7 & 84.24 & 84.47 \\
LISA & 84.72 &	84.57	& 84.64 \\ 
\ \ \ \ +D\&M & {\bf 86.02} &	{\bf 86.05} &	{\bf 86.04}  \\
& & & \\
Brown Test &  P & R & F1 \\ \hline \hline
\citet{he2018jointly} & 74.2 & 73.1 & 73.7 \\
\citet{tan2018deep} & 73.5 & 74.6 & 74.1 \\ \hline
SA & 73.89 & 72.39 & 73.13 \\
LISA & 74.77 & 74.32 &	74.55 \\ 
\ \ \ \ +D\&M & {\bf 76.65} & {\bf 76.44} & {\bf 76.54} \\ 
\end{tabular}
\caption{Precision, recall and F1 on CoNLL-2005 with gold predicates. \label{tab:conll05-gold-pred}}
\end{table}

\subsection{Semantic role labeling \label{sec:conll05}}

Table~\ref{tab:conll05-results} lists precision, recall and F1 on the CoNLL-2005 development and test sets using predicted predicates. For models using GloVe embeddings, our syntax-free SA model already achieves a new state-of-the-art by jointly predicting predicates, POS and SRL. LISA with its own parses performs comparably to SA, but when supplied with D\&M parses LISA out-performs the previous state-of-the-art by 2.5 F1 points. On the out-of-domain Brown test set, LISA also performs comparably to its syntax-free counterpart with its own parses, but with D\&M parses LISA performs exceptionally well, more than 3.5 F1 points higher than \citet{he2018jointly}. Incorporating ELMo embeddings improves all scores. The gap in SRL F1 between models using LISA and D\&M parses is smaller due to LISA's improved parsing accuracy (see \S\ref{sec:parse-pos-results}), but LISA with D\&M parses still achieves the highest F1: nearly 1.0 absolute F1 higher than the previous state-of-the art on WSJ, and more than 2.0 F1 higher on Brown. In both settings LISA leverages domain-agnostic syntactic information rather than over-fitting to the newswire training data which leads to high performance even on out-of-domain text.

To compare to more prior work we also evaluate our models in the artificial setting where gold predicates are provided at test time. For fair comparison we use GloVe embeddings, provide predicate indicator embeddings on the input and re-encode the sequence relative to each gold predicate. Here LISA still excels: with D\&M parses, LISA out-performs the previous state-of-the-art by more than 2 F1 on both WSJ and Brown. 

Table~\ref{tab:conll12-results} reports precision, recall and F1 on the CoNLL-2012 test set. We observe performance similar to that observed on ConLL-2005: Using GloVe embeddings our SA baseline already out-performs \citet{he2018jointly} by nearly 1.5 F1. With its own parses, LISA slightly under-performs our syntax-free model, but when provided with stronger D\&M parses LISA out-performs the state-of-the-art by more than 2.5 F1. Like CoNLL-2005, ELMo representations improve all models and close the F1 gap between models supplied with LISA and D\&M parses. On this dataset ELMo also substantially narrows the difference between models with- and without syntactic information. This suggests that for this challenging dataset, ELMo already encodes much of the information available in the D\&M parses. Yet, higher accuracy parses could still yield improvements since providing gold parses increases F1 by 4 points even with ELMo embeddings.

\begin{table}
\begin{tabular}{llll}
Dev & P & R & F1 \\ \hline \hline
\multicolumn{4}{c}{GloVe} \\ \hline
\citet{he2018jointly} & 79.2 & 79.7 & 79.4 \\ \hline
SA & 82.32 & 79.76 & 81.02 \\ 
LISA & 81.77 & 79.65 & 80.70 \\
\ \ \ \ +D\&M & {\bf 82.97} & {\bf 81.14} &	{\bf 82.05} \\
\ \ \ \ \emph{+Gold} & \emph{87.57} & \emph{85.32} & \emph{86.43} \\ 
\hline
 & & & \\
\multicolumn{4}{c}{ELMo} \\ \hline
\citet{he2018jointly} & 82.1 & {\bf 84.0} & 83.0 \\ \hline
SA & 84.35 & 82.14 & 83.23 \\ 
LISA & {\bf 84.19} & 82.56 & {\bf 83.37} \\
\ \ \ \ +D\&M & 84.09 & 82.65 & 83.36 \\
\ \ \ \ \emph{+Gold} & \emph{88.22} & \emph{86.53} & \emph{87.36} \\

 & & & \\
Test &  P & R & F1 \\ \hline \hline
\multicolumn{4}{c}{GloVe} \\ \hline
\citet{he2018jointly} & 79.4 & 80.1 & 79.8 \\ \hline
SA & 82.55 & 80.02 & 81.26 \\ 
LISA &  81.86 &	79.56 &	80.70 \\ 
\ \ \ \ +D\&M & {\bf 83.3} & {\bf 81.38} &	{\bf 82.33} \\ 
\hline
 & & & \\
\multicolumn{4}{c}{ELMo} \\ \hline
\citet{he2018jointly} & 81.9 & {\bf 84.0} & 82.9 \\ \hline
SA & {\bf 84.39} & 82.21 & 83.28 \\ 
LISA & 83.97 & 82.29 & 83.12 \\
\ \ \ \ +D\&M & 84.14 & 82.64 & {\bf 83.38} \\	

\end{tabular}
\caption{Precision, recall and F1 on the CoNLL-2012 development and test sets. Italics indicate a synthetic upper bound obtained by providing a gold parse at test time.\label{tab:conll12-results}}
\end{table}


\subsection{Parsing, POS and predicate detection \label{sec:parse-pos-results}}

\begin{table}
\begin{tabular}{llrrr} 		
Data & Model & POS & UAS & LAS \\ \hline \hline
\multirow{3}{*}{WSJ} & D\&M$_{E}$ & --- & 96.48 & 94.40 \\
& LISA$_{G}$ & 96.92 & 94.92 & 91.87 \\ 
& LISA$_{E}$ & 97.80 & 96.28 & 93.65 \\ \hline
\multirow{3}{*}{Brown} & D\&M$_{E}$ & --- & 92.56 & 88.52 \\
& LISA$_{G}$ & 94.26 & 90.31 & 85.82 \\ 
& LISA$_{E}$ & 95.77 & 93.36 & 88.75 \\ \hline
\multirow{3}{*}{CoNLL-12} & D\&M$_{E}$ & --- & 94.99 & 92.59 \\
& LISA$_{G}$ & 96.81 & 93.35 & 90.42 \\
& LISA$_{E}$ & 98.11 & 94.84 & 92.23 \\
\end{tabular}
\caption{\label{parsing-numbers} Parsing (labeled and unlabeled attachment) and POS accuracies attained by the models used in SRL experiments on test datasets. Subscript $G$ denotes GloVe and $E$ ELMo embeddings.}
\end{table}

We first report the labeled and unlabeled attachment scores (LAS, UAS) of our parsing models on the CoNLL-2005 and 2012 test sets (Table~\ref{parsing-numbers}) with GloVe ($G$) and ELMo ($E$) embeddings. D\&M achieves the best scores. Still, LISA's GloVe UAS is comparable to popular off-the-shelf dependency parsers such as spaCy,\footnote{spaCy reports 94.48 UAS on WSJ using Stanford dependencies v3.3: \protect\url{https://spacy.io/usage/facts-figures}} and with ELMo embeddings comparable to the standalone D\&M parser. The difference in parse accuracy between LISA$_G$ and D\&M likely explains the large increase in SRL performance we see from decoding with D\&M parses in that setting.

\begin{table}
\setlength{\tabcolsep}{5pt}
\begin{tabular}{lllllll}


& Model & P & R & F1 \\ \hline \hline
\multirow{2}{*}{WSJ} & \citet{he2017deep} & 94.5 & 98.5 & 96.4  \\
& LISA & 98.9 &  97.9 & 98.4 \\ \hline

\multirow{2}{*}{Brown} & \citet{he2017deep} & 89.3 & 95.7 & 92.4 \\ 
& LISA & 95.5 &  91.9 &  93.7 \\ \hline

CoNLL-12 & LISA & 99.8 & 94.7 &	97.2 \\ 
\end{tabular}
\caption{Predicate detection precision, recall and F1 on CoNLL-2005 and CoNLL-2012 test sets. \label{tab:preds}}
\end{table}

In Table~\ref{tab:preds} we present predicate detection precision, recall and F1 on the CoNLL-2005 and 2012 test sets. SA and LISA with and without ELMo attain comparable scores so we report only LISA+GloVe. We compare to \citet{he2017deep} on CoNLL-2005, the only cited work reporting comparable predicate detection F1. LISA attains high predicate detection scores, above 97 F1, on both in-domain datasets, and out-performs \citet{he2017deep} by 1.5-2 F1 points even on the out-of-domain Brown test set, suggesting that multi-task learning works well for SRL predicate detection. 

\begin{table}

\begin{tabular}{lllll}
& L+/D+ & L--/D+ & L+/D-- & L--/D-- \\ \hline \hline
Proportion & 26\% &	12\% &	4\% &	56\% \\ \hline
SA & 79.29 & 75.14	& 75.97 &	75.08 \\ 
LISA & 79.51 &	74.33 &	79.69 &	75.00 \\
\ \ \ \ +D\&M & 79.03 &	76.96 &	77.73 &	76.52 \\
\ \ \ \ \emph{+Gold} & \emph{79.61} & \emph{78.38} & \emph{81.41} & \emph{80.47} \\
\end{tabular}
\caption{Average SRL F1 on CoNLL-2005 for sentences where LISA (L) and D\&M (D) parses were completely correct (+) or incorrect (--). \label{tab:parse-srl-by-sents}}
\end{table}

\subsection{Analysis \label{sec:analysis}}


First we assess SRL F1 on sentences divided by parse accuracy. Table \ref{tab:parse-srl-by-sents} lists average SRL F1 (across sentences) for the four conditions of LISA and D\&M parses being correct or not ({\bf L$\pm$}, {\bf D$\pm$}). Both parsers are correct on 26\% of sentences. Here there is little difference between any of the models, with LISA models tending to perform slightly better than SA. Both parsers make mistakes on the majority of sentences (57\%), difficult sentences where SA also performs the worst. These examples are likely where gold and D\&M parses improve the most over other models in overall F1: Though both parsers fail to correctly parse the entire sentence, the D\&M parser is less wrong (87.5 vs. 85.7 average LAS), leading to higher SRL F1 by about 1.5 average F1.

Following \citet{he2017deep}, we next apply a series of corrections to model predictions in order to understand which error types the gold parse resolves: e.g. \emph{Fix Labels} fixes labels on spans matching gold boundaries, and \emph{Merge Spans} merges adjacent predicted spans into a gold span.\footnote{Refer to \citet{he2017deep} for a detailed explanation of the different error types.}

\begin{figure}
\includegraphics[scale=0.52]{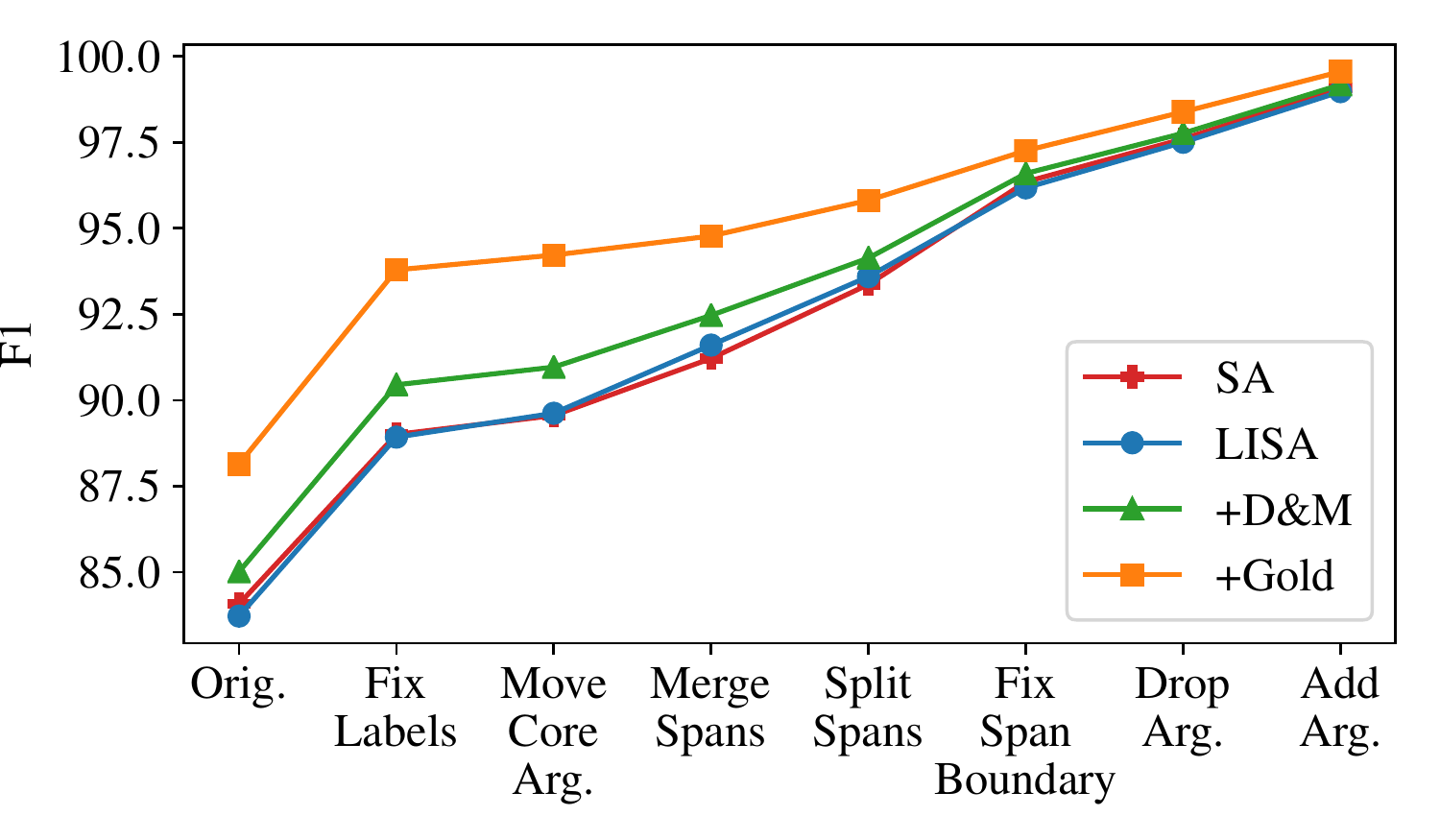}
\caption{Performance of CoNLL-2005 models after performing corrections from \citet{he2017deep}. \label{errors-fig}}
\end{figure}

\begin{figure}
\centering
\includegraphics[scale=0.55]{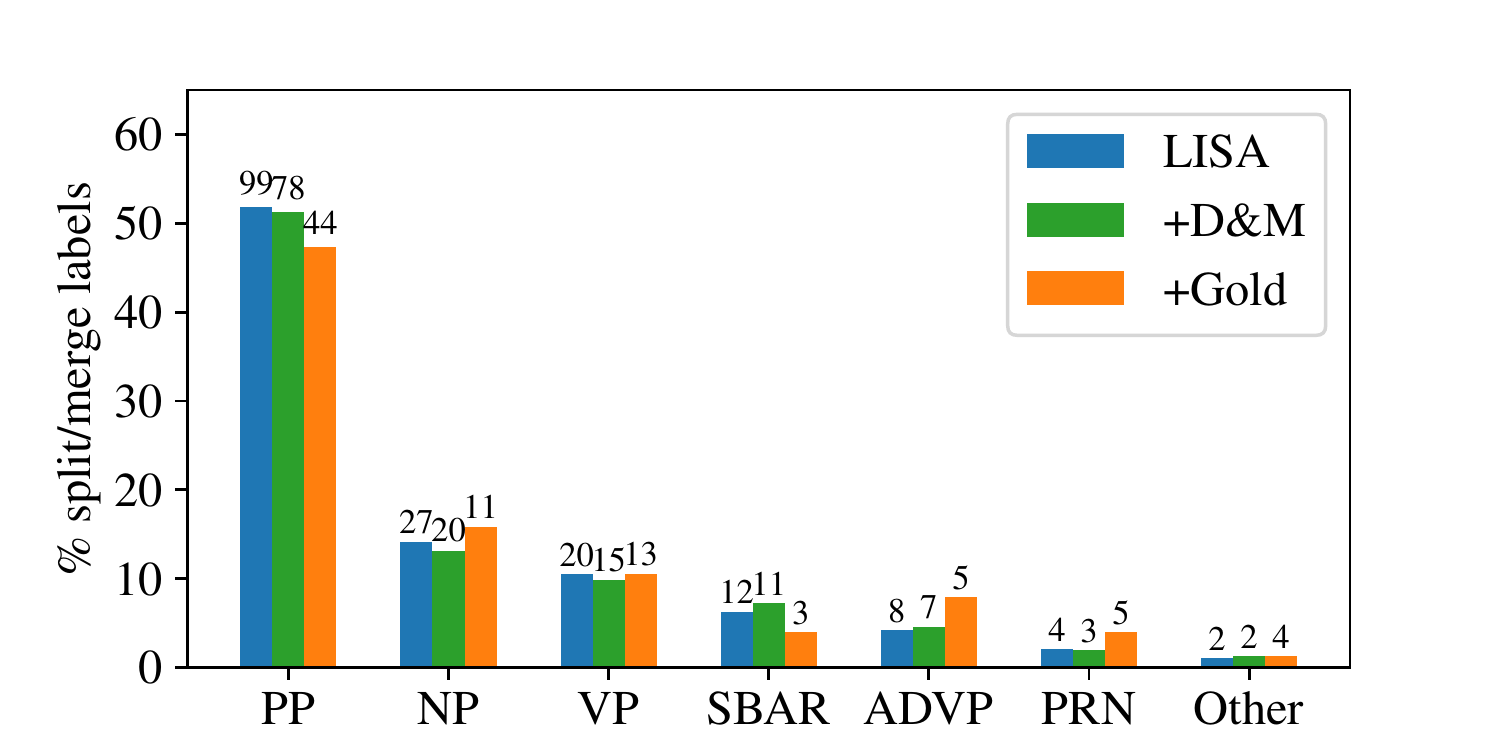}
\caption{Percent and count of split/merge corrections performed in Figure \ref{errors-fig}, by phrase type. \label{fig:phrase-bar}}
\end{figure}

In Figure \ref{errors-fig} we see that much of the performance gap between the gold and predicted parses is due to span boundary errors (\emph{Merge Spans}, \emph{Split Spans} and \emph{Fix Span Boundary}), which supports the hypothesis proposed by \citet{he2017deep} that incorporating syntax could be particularly helpful for resolving these errors. 
\citet{he2017deep} also point out that these errors are due mainly to prepositional phrase (PP) attachment mistakes. We also find this to be the case: Figure \ref{fig:phrase-bar} shows a breakdown of split/merge corrections by phrase type. Though the number of corrections decreases substantially across phrase types, the proportion of corrections attributed to PPs remains the same (approx. 50\%) even after providing the correct PP attachment to the model, indicating that PP span boundary mistakes are a fundamental difficulty for SRL.

\section{Conclusion}
We present linguistically-informed self-attention: a multi-task neural network model that effectively incorporates rich linguistic information for semantic role labeling. LISA out-performs the state-of-the-art on two benchmark SRL datasets, including out-of-domain. Future work will explore improving LISA's parsing accuracy, developing better training techniques and adapting to more tasks.

\section*{Acknowledgments}
We are grateful to Luheng He for helpful discussions and code, Timothy Dozat for sharing his code, and to the NLP reading groups at Google and UMass and the anonymous reviewers for feedback on drafts of this work. This work was supported in part by an IBM PhD Fellowship Award to E.S., in part by the Center for Intelligent Information Retrieval, and in part by the National Science Foundation under Grant Nos. DMR-1534431 and IIS-1514053. Any opinions, findings, conclusions or recommendations expressed in this material are those of the authors and do not necessarily reflect those of the sponsor.

\bibliography{emnlp2018}
\bibliographystyle{acl_natbib_nourl}

\clearpage\newpage
\appendix
\section{Supplemental Material \label{sec:supplemental}}

\subsection{Supplemental analysis \label{app:analysis}}
Here we continue the analysis from \S\ref{sec:analysis}. All experiments in this section are performed on CoNLL-2005 development data unless stated otherwise.

\begin{table}
\begin{tabular}{llll}
CoNLL-2005 & Greedy F1 & Viterbi F1 & $\Delta$ F1 \\ \hline \hline
LISA & 81.99 & 82.24 & +0.25 \\
\ \ \ \ +D\&M & 83.37 & 83.58 & +0.21 \\
\ \ \ \ \emph{+Gold} & \emph{86.57} &	\emph{86.81} &	\emph{+0.24} \\
& & & \\
CoNLL-2012 & Greedy F1 & Viterbi F1 & $\Delta$ F1 \\ \hline \hline
LISA & 80.11	& 80.70	 & +0.59 \\
\ \ \ \ +D\&M & 81.55 &	82.05 & +0.50 \\
\ \ \ \ \emph{+Gold} & \emph{85.94} &	\emph{86.43} &	\emph{+0.49} \\
\end{tabular}

\caption{Comparison of development F1 scores with and without Viterbi decoding at test time. \label{viterbi-table}}
\end{table}

First, we compare the impact of Viterbi decoding with LISA, D\&M, and gold syntax trees (Table \ref{viterbi-table}), finding the same trends across both datasets. We find that Viterbi has nearly the same impact for LISA, D\&M and gold parses: Gold parses provide little improvement over predicted parses in terms of BIO label consistency. 

\begin{figure}
\includegraphics[scale=0.52]{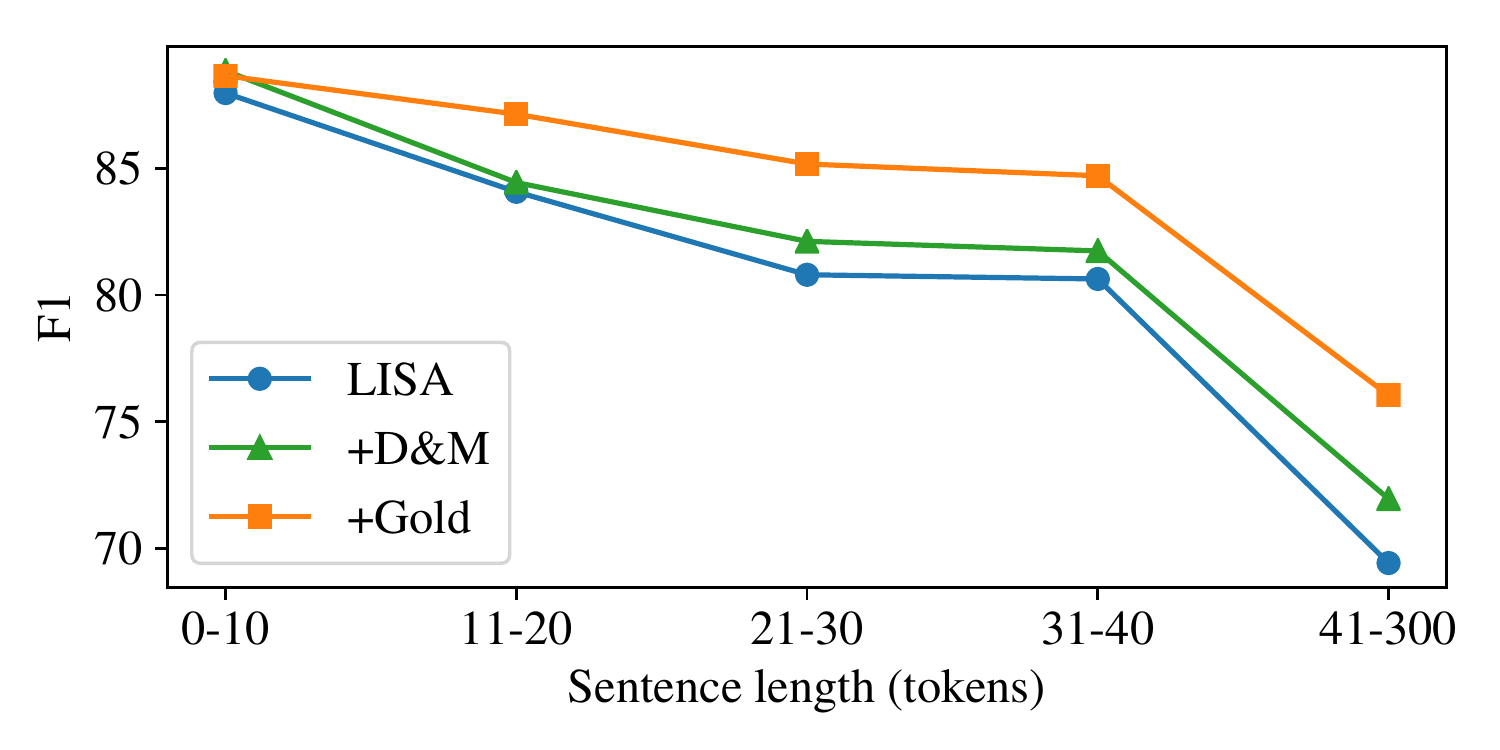}
\caption{F1 score as a function of sentence length.\label{fig:length}}
\end{figure}

\begin{figure}
\includegraphics[scale=0.52]{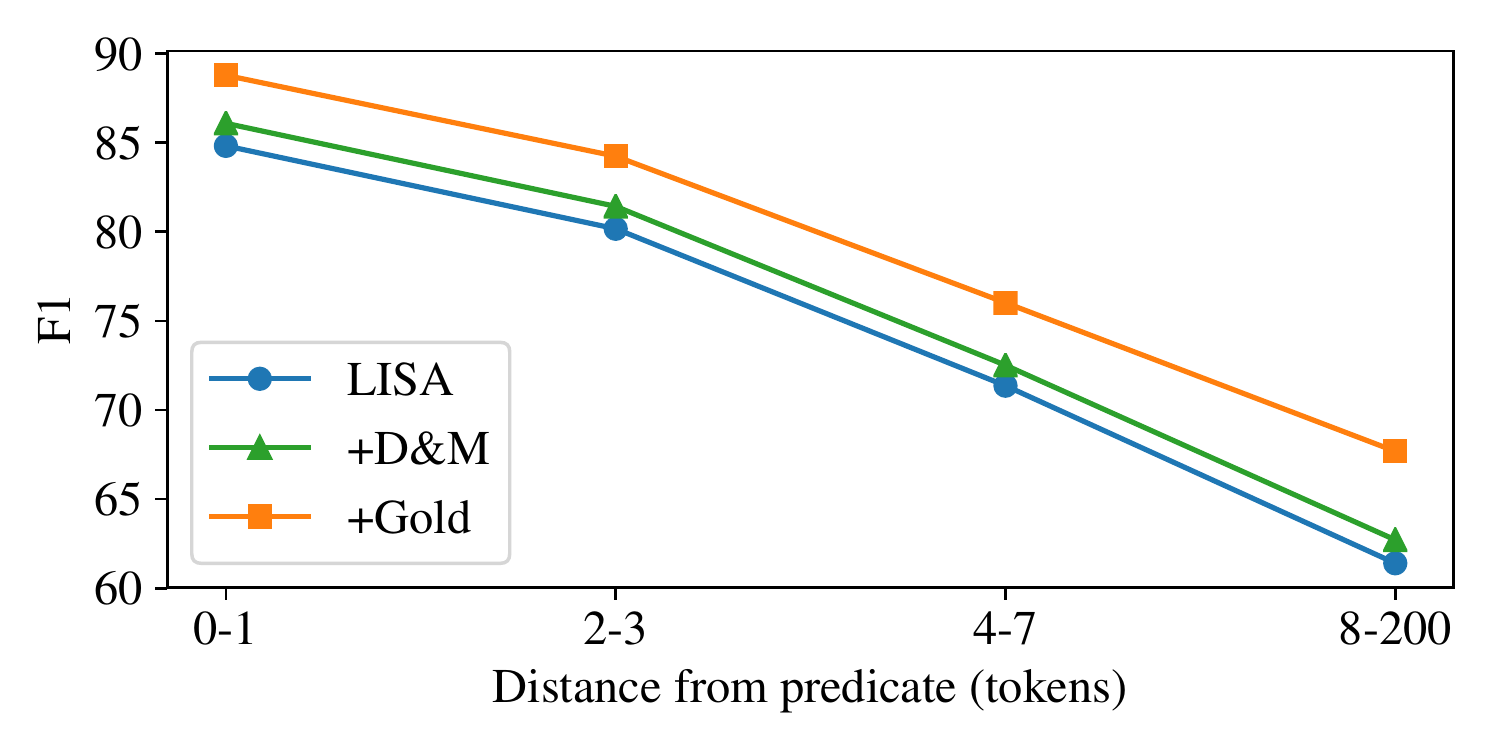}
\caption{CoNLL-2005 F1 score as a function of the distance of the predicate from the argument span.\label{fig:dist}}
\end{figure}

We also assess SRL F1 as a function of sentence length and distance from span to predicate. In Figure \ref{fig:length} we see that providing LISA with gold parses is particularly helpful for sentences longer than 10 tokens. This likely directly follows from the tendency of syntactic parsers to perform worse on longer sentences. With respect to distance between arguments and predicates, (Figure \ref{fig:dist}), we do not observe this same trend, with all distances performing better with better parses, and especially gold. 



\begin{table}
\begin{tabular}{lllll}
& L+/D+ & L-/D+ & L+/D- & L-/D- \\ \hline \hline
Proportion & 37\% &	10\% &	4\% &	49\% \\ \hline
SA & 76.12 & 75.97 & 82.25 &	65.78 \\ 
LISA & 76.37 &	72.38 &	85.50 &	65.10 \\
\ \ \ \ +D\&M & 76.33	& 79.65 &	75.62 &	66.55 \\
\ \ \ \ \emph{+Gold} & \emph{76.71} & \emph{80.67} & \emph{86.03} & \emph{72.22} \\
\end{tabular}
\caption{Average SRL F1 on CoNLL-2012 for sentences where LISA (L) and D\&M (D) parses were correct (+) or incorrect (-). \label{tab:app:parse-srl-by-sents}}
\end{table}

\subsection{Supplemental results}

Due to space constraints in the main paper we list additional experimental results here. Table \ref{tab:conll05-gold-pred-dev} lists development scores on the CoNLL-2005 dataset with predicted predicates, which follow the same trends as the test data.

\begin{table}
\begin{tabular}{llll}
WSJ Dev & P & R & F1 \\ \hline \hline
\citet{he2018jointly} & 84.2 & 83.7 & 83.9 \\
\citet{tan2018deep} & 82.6	& 83.6 &	83.1 \\ \hline
SA & 83.12 &	82.81 &	82.97 \\
LISA & 83.6 &	83.74	& 83.67 \\
\ \ \ \ +D\&M & {\bf 85.04} &	{\bf 85.51} &	{\bf 85.27} \\
\ \ \ \ \emph{+Gold} & \emph{89.11} &	\emph{89.38} & 	\emph{89.25}
\end{tabular}
\caption{Precision, recall and F1 on the CoNLL-2005 development set with gold predicates. \label{tab:conll05-gold-pred-dev}}
\end{table}

\subsection{Data and pre-processing details}

We initialize word embeddings with 100d pre-trained GloVe embeddings trained on 6 billion tokens of Wikipedia and Gigaword \citep{pennington2014glove}. We evaluate the SRL performance of our models using the \texttt{srl-eval.pl} script provided by the CoNLL-2005 shared task,\footnote{\protect\url{http://www.lsi.upc.es/~srlconll/srl-eval.pl}} which computes segment-level precision, recall and F1 score. We also report the predicate detection scores output by this script. We evaluate parsing using the \texttt{eval.pl} CoNLL script, which excludes punctuation.

We train distinct D\&M parsers for CoNLL-2005 and CoNLL-2012. Our D\&M parsers are trained and validated using the same SRL data splits, except that for CoNLL-2005 section 22 is used for development (rather than 24), as this section is typically used for validation in PTB parsing. We use Stanford dependencies v3.5 \citep{deMarneffe2008} and POS tags from the Stanford CoreNLP \texttt{left3words} model \citep{toutanova2003feature}. We use the pre-trained ELMo models\footnote{\protect\url{https://github.com/allenai/bilm-tf}} and learn task-specific combinations of the ELMo representations which are provided as input instead of GloVe embeddings to the D\&M parser with otherwise default settings.

\subsubsection{CoNLL-2012}
We follow the CoNLL-2012 split used by \citet{he2018jointly} to evaluate our models, which uses the annotations from here\footnote{\protect\url{http://cemantix.org/data/ontonotes.html}} but the subset of those documents from the CoNLL-2012 co-reference split described here\footnote{\protect\url{http://conll.cemantix.org/2012/data.html}} \citep{pradhan2013towards}. This dataset is drawn from seven domains: newswire, web, broadcast news and conversation, magazines, telephone conversations, and text from the bible. The text is annotated with gold part-of-speech, syntactic constituencies, named entities, word sense, speaker, co-reference and semantic role labels based on the PropBank guidelines \citep{palmer2005proposition}. Propositions may be verbal or nominal, and there are 41 distinct semantic role labels, excluding continuation roles and including the predicate. We convert the semantic proposition and role segmentations to BIO boundary-encoded tags, resulting in 129 distinct BIO-encoded tags (including continuation roles). 

\subsubsection{CoNLL-2005}
The CoNLL-2005 data \citep{carreras2005introduction} is based on the original PropBank corpus \citep{palmer2005proposition}, which labels the Wall Street Journal portion of the Penn TreeBank corpus (PTB) \citep{marcus1993building} with predicate-argument structures, plus a challenging out-of-domain test set derived from the Brown corpus \citep{francis1964manual}. This dataset contains only verbal predicates, though some are multi-word verbs, and 28 distinct role label types. We obtain 105 SRL labels including continuations after encoding predicate argument segment boundaries with BIO tags.

\subsection{Optimization and hyperparameters}
We train the model using the Nadam \citep{dozat2016incorporating} algorithm for adaptive stochastic gradient descent (SGD), which combines Adam \citep{kingma2014adam} SGD with Nesterov momentum \citep{nesterov1983method}. We additionally vary the learning rate $lr$ as a function of an initial learning rate $lr_0$ and the current training step $step$ as described in \citet{vaswani2017attention} using the following function:
\begin{align}
lr = lr_0 \cdot \min(step^{-0.5},  step\cdot warm^{-1.5})
\end{align}
which increases the learning rate linearly for the first $warm$ training steps, then decays it proportionally to the inverse square root of the step number. We found this learning rate schedule essential for training the self-attention model. We only update optimization moving-average accumulators for parameters which receive gradient updates at a given step.\footnote{Also known as \emph{lazy} or \emph{sparse} optimizer updates.}

In all of our experiments we used initial learning rate 0.04, $\beta_1=0.9$, $\beta_2=0.98$, $\epsilon=1\times10^{-12}$ and dropout rates of 0.1 everywhere. We use 10 or 12 self-attention layers made up of 8 attention heads each with embedding dimension 25, with 800d feed-forward projections. In the syntactically-informed attention head, $Q_{parse}$ has dimension 500 and $K_{parse}$ has dimension 100. The size of $predicate$ and $role$ representations and the representation used for joint part-of-speech/predicate classification is 200. We train with $warm=8000$ warmup steps and clip gradient norms to 1. We use batches of approximately 5000 tokens.

\end{document}